\documentclass[lettersize,journal]{IEEEtran}
\usepackage{amsmath,amsfonts}
\usepackage{algorithmic}
\usepackage{algorithm}
\usepackage{array}
\usepackage[caption=false,font=normalsize,labelfont=sf,textfont=sf]{subfig}
\usepackage{textcomp}
\usepackage{stfloats}
\usepackage{url}
\usepackage{verbatim}
\usepackage{graphicx}
\usepackage{cite}
\usepackage{tabularx}
\usepackage{scalerel}
\usepackage{tikz}
\usetikzlibrary{svg.path}
\definecolor{orcidlogocol}{HTML}{A6CE39}
\tikzset{
  orcidlogo/.pic={
    \fill[orcidlogocol] svg{M256,128c0,70.7-57.3,128-128,128C57.3,256,0,198.7,0,128C0,57.3,57.3,0,128,0C198.7,0,256,57.3,256,128z};
    \fill[white] svg{M86.3,186.2H70.9V79.1h15.4v48.4V186.2z}
                 svg{M108.9,79.1h41.6c39.6,0,57,28.3,57,53.6c0,27.5-21.5,53.6-56.8,53.6h-41.8V79.1z M124.3,172.4h24.5c34.9,0,42.9-26.5,42.9-39.7c0-21.5-13.7-39.7-43.7-39.7h-23.7V172.4z}
                 svg{M88.7,56.8c0,5.5-4.5,10.1-10.1,10.1c-5.6,0-10.1-4.6-10.1-10.1c0-5.6,4.5-10.1,10.1-10.1C84.2,46.7,88.7,51.3,88.7,56.8z};
  }
}

\newcommand\orcidicon[1]{\href{https://orcid.org/#1}{\mbox{\scalerel*{
\begin{tikzpicture}[yscale=-1,transform shape]
\pic{orcidlogo};
\end{tikzpicture}
}{|}}}}

\usepackage{hyperref} 

\newcommand\copyrighttext{%
  \footnotesize \textcopyright 2022 IEEE. Personal use of this material is permitted. Permission
from IEEE must be obtained for all other uses, in any current or future
media, including reprinting/republishing this material for advertising or
promotional purposes, creating new collective works, for resale or
redistribution to servers or lists, or reuse of any copyrighted
component of this work in other works.}
\newcommand\copyrightnotice{%
\begin{tikzpicture}[remember picture,overlay]
\node[anchor=south,yshift=7pt] at (current page.south) {\fbox{\parbox{\dimexpr\textwidth-\fboxsep-\fboxrule\relax}{\copyrighttext}}};
\end{tikzpicture}%
}

\begin{document}

\title{Neural Combinatorial Optimization: a New Player in the Field}

\author{Andoni I. Garmendia\orcidicon{0000-0002-7243-6116}, Josu Ceberio\orcidicon{0000-0001-7120-6338}, Alexander Mendiburu\orcidicon{0000-0002-7271-1931},~\IEEEmembership{Member,~IEEE,}
}



\maketitle
\copyrightnotice

\begin{abstract}
Neural Combinatorial Optimization attempts to learn good heuristics for solving a set of problems using Neural Network models and Reinforcement Learning. Recently, its good performance has encouraged many practitioners to develop neural architectures for a wide variety of combinatorial problems. However, the incorporation of such algorithms in the conventional optimization framework has raised many questions related to their performance and the experimental comparison with other methods such as exact algorithms, heuristics and metaheuristics. 
This paper presents a critical analysis on the incorporation of algorithms based on neural networks into the classical combinatorial optimization framework.
Subsequently, a comprehensive study is carried out to analyse the fundamental aspects of such algorithms, including \textit{performance}, \textit{transferability}, \textit{computational cost} and \textit{generalization to larger-sized instances}.
To that end, we select the Linear Ordering Problem as a case of study, an NP-hard problem, and develop a Neural Combinatorial Optimization model to optimize it. 
Finally, we discuss how the analysed aspects apply to a general learning framework, and suggest new directions for future work in the area of Neural Combinatorial Optimization algorithms.

\end{abstract}


\section{Introduction}
\IEEEPARstart{C}{ombinatorial} optimization deals with problems with an objective function that has to be maximized over a set of combinatorial alternatives \cite{Korte2011}. The most naive way of obtaining the optimal solution is to list all the feasible solutions, evaluating them using the objective function and selecting the optimal one. Nevertheless, a brute-force approach lacks practicality when the size of the problem is too large, as the time needed to solve it grows exponentially with the problem size. 

Combinatorial optimization problems (COPs) have been traditionally approached either by exact methods or heuristic methods. Exact methods guarantee an optimal solution but, as their algorithmic complexity is high, exact approaches become useless for medium to large problem sizes. Conversely, heuristic methods do not guarantee to reach the optimal solution, but they try to reach a good solution in a reasonable time budget. The effectiveness of a heuristic method depends on its ability to identify and exploit the relevant information of the problem at hand, in order to optimize it efficiently. In this line, as a generalization of heuristic algorithms, metaheuristics (MH) introduce higher level generic procedures that guide the search process, making them "general-purpose methods" \cite{Blum2003}.

Even though a lot of work has been done in the development of MH algorithms, in the last few decades the research in that area has reached a point of maturity, with hardly any considerable improvements being made. In contrast, Deep Learning (DL) methods have recently entered the optimization paradigm, introducing a breath of fresh air. The recent success of DL in fields such as machine translation \cite{Cho2014}, biology \cite{Jumper2021} and board games \cite{Schrittwieser2020}, has not only attracted the attention of many machine learning practitioners into the optimization field, but has also captured the interest of the optimization community about the possibilities of DL. 

The main building blocks of DL algorithms, Neural Networks (NN), were proposed in the 80s to solve COPs as a promising component to incorporate the relevant information of a problem by learning \cite{Hopfield1985}. However, only in the last few decades have NNs become an interesting alternative, due to the increase in computational capacity and the development of complex NN models, which enable the design of competitive algorithms. Recent reviews \cite{Bengio2021, Talbi2021} present two taxonomies that differentiate the ways and the stages in which DL can be applied to COPs. In general, two main groups can be distinguished:  end-to-end methods and hybrid methods. \textbf{End-to-end methods} are designed to solve a particular problem by means of NN models without needing any additional algorithm. Regarding \textbf{hybrid methods}, they combine DL models with previously mentioned exact and heuristic methods. According to \cite{Bengio2021, Talbi2021}, hybrid methods can be split into two groups: 
\begin{enumerate}
  \item \textbf{Learning to configure algorithms}. Almost all the metaheuristics that have been proposed in the literature have a set of configurable parameters and operators that must be tuned carefully for the algorithm to perform efficiently. This parameter tuning has concerned the community for years \cite{Yu2020}, and different proposals have been developed using DL to learn how to configure those parameters \cite{Huang2019}.
  \item \textbf{DL alongside optimization algorithms}. This group includes two-level algorithms, in which a heuristic calls a DL model to assist in some decisions, such as the most appropriate crossover or mutation operator, or selecting the most promising branching variable. For instance, a NN can be used to select the next branch and improve the exploration of a Monte Carlo Tree Search algorithm \cite{Xing2020}.
\end{enumerate}

Though there is much yet to come, hybrid methods generally recreate previously studied techniques with the addition of DL modules \cite{Talbi2021}. Conversely, end-to-end proposals work on the basis of an NN model as the core of the algorithm, and contrary to the conventional methods, the optimization process consists of 2 steps: a learning phase in which the model will be trained, and the inference, in which the model gives a solution to a new problem instance. Therefore, this approach is significantly different to that used by traditional optimization algorithms and, even if interesting proposals have been presented in the literature, the benefits and limitations of end-to-end proposals have not been studied in-depth. Thus, in this paper a discussion about end-to-end models is opened, focusing on the most relevant aspects: (1) \textbf{Performance}. How good are the solutions provided by these models? Are they competitive with the state-of-the-art methods? (2) \textbf{Training data \& Transferability}. Which type of data is required to train the model? Is it necessary to have hundreds or even thousands of instances of the real world problem to train the model? Can random instances be used instead? (3) \textbf{Computational cost}. Considering the computation resources and time required to train these NN-based models, are they affordable for medium to large size problems? (4) \textbf{Generalization to large size instances}. Related to the previous issue, in the case when a large enough model can not be trained due to a memory limitation, is it possible to competitively apply a model trained for small size instances to larger instances? Moreover, can a model trained for a problem be successfully applied to another problem?

In addition to analysing these aspects on the works presented in the literature, with illustrative purposes, we take a practical case study to apply the analysis over end-to-end models. Particularly, we develop an end-to-end model to solve the Linear Ordering Problem (LOP) \cite{Ceberio2015}, a well-known NP-hard COP. Our purpose is not to present a state-of-the-art algorithm, but to guide the reader during the process of implementation and evaluation of the method and address the aforementioned aspects. Moreover, we conduct an experimentation of its capabilities with a broad comparison, using a diverse set of conventional methods, including an exact solver \cite{Achterberg2009}, a constructive heuristic \cite{Becker1967} and two state-of-the-art metaheuristics \cite{Lugo2021,Santucci2020}. Conducted experiments show that the end-to-end model is competitive against the constructive heuristic, although it falls behind the metaheuristic algorithms. However, the NN model is capable of generalizing the learnt knowledge to larger instances and transfer it to other types of instances. Not limited to that, a number of promising research lines are described for future investigations on the application of end-to-end models over COPs.

The rest of the paper is organized as follows. Section \ref{related_work} introduces a revision of meaningful works in the neural end-to-end framework. Section \ref{analysis} presents four features of the critical analysis that arise from the introduction of DL algorithms to the optimization field. We try to find answers to a number of relevant questions on the basis of a case of study, the Linear Ordering Problem, which is defined in Section \ref{problem_definition}, and an end-to-end approach, presented in Section \ref{architecture}. A broad set of experiments is conducted in Section \ref{experiments}. Obtained results are discussed in Section \ref{discussion}, where we also suggest new directions for future work in the area of learning algorithms. Finally, Section \ref{conclusion} concludes the paper.

\section{Related Work}
\label{related_work}


As stated previously, end-to-end models for combinatorial optimization problems are the scope of this paper. In that sense, this section analyses an exhaustive set of this type of proposals, highlighting for each case the methodological and practical advances introduced.

In general terms, the optimization process of end-to-end models follows a training-inference pipeline. In the training phase, a set of instances is used to learn the parameters of the NN model. In this step, two main learning scenarios arise: Supervised Learning (SL) and Reinforcement Learning (RL). 

\begin{figure}
    \centering
    \includegraphics[width=0.45\textwidth]{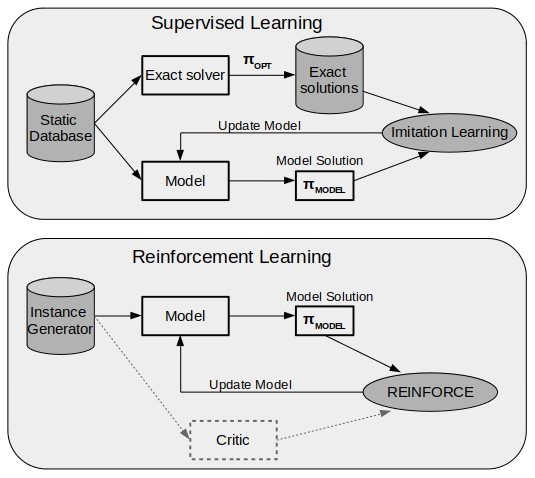}
    \caption{General structure of different learning methods. In \textbf{Supervised Learning}, the model imitates an exact (or good enough) solver. Usually, a static dataset of pre-solved instances is needed, where the NN model takes the labels for the training. In \textbf{Reinforcement Learning}, the NN model learns based on the reward obtained from the inferred solution. Regarding the learning data, the use of random generators is common in RL methods.}
    \label{fig:nco}
\end{figure}

In SL, the NN model learns to imitate an optimal (or good) policy \cite{Gasse2019}. The NN model is fed with a collection of solved instances that have been previously computed by an exact or an approximate solver (see SL in Fig. \ref{fig:nco}). An example of this approach can be seen in \cite{Vinyals2015}, where the authors propose an end-to-end algorithm to solve the Travelling Salesman Problem\footnote{Among the most popular combinatorial problems, the Travelling Salesman Problem (TSP) has been one of the most studied problems. The goal in TSP is to find the shortest route between a set of cities that must be visited just once.}. The proposed algorithm embeds the information of the instance by an NN model and constructs a solution for the problem iteratively, adding a city to the solution at a time. Under this approach, Vinyals et al. \cite{Vinyals2015} use an architecture called Pointer Network, a sequence-to-sequence model that points to one of the input cities. 

However, using SL has some serious drawbacks, since obtaining a large set of labels is not usually tractable and affects the scalability of the method \cite{Bengio2021}. Moreover, imitation learning may fail to abstract the problem knowledge when the imitated policy is suboptimal or there are multiple optimal solutions. Instead, RL proposes a more suitable procedure for solving COPs, where an agent learns how to act, without supervision, based on the rewards it receives through its history, i.e., by experience \cite{Sutton2018} (see RL in Fig. \ref{fig:nco}). In this learning scenario, Bello et al. \cite{Bello2016} introduced a framework called Neural Combinatorial Optimization (NCO), which uses RL and NN models to learn approximate solutions for (a set of) combinatorial problems in an end-to-end manner. In that paper, the authors improve the performance of the model presented by Vinyals et al. \cite{Vinyals2015}, using a similar architecture but replacing SL with RL. Moreover, Bello et al. \cite{Bello2016} achieved better results than a classical heuristic (Christofides \cite{Christofides1976}) and  OR-Tools’ local search algorithm \cite{Google2016}. 

Motivated by the good results obtained by Bello et al. \cite{Bello2016}, most of the papers dealing with end-to-end models have followed a similar learning scheme, and the scientific advances have focused on changes in the way the solutions are inferred. In this context, two main strategies are present in the literature: (1) \textbf{one-shot} methods predict a probability matrix that describes the probability of an item belonging to a specific part of the solution. The model infers the matrix in one go and solutions are then obtained by performing a beam-search over the probability matrix\cite{Joshi2019}. (2) \textbf{Autoregressive} methods construct the solutions sequentially, adding one item to the solution in each iteration and updating its current state\cite{Joshi2020}.

In addition to the learning scenarios, recent works have also performed changes in the NN model architectures. As an improvement of Pointer Networks \cite{Vinyals2015, Bello2016}, Graph Neural Networks (GNN) \cite{Cappart2021} address the limitation of having an order invariant input, i.e., GNNs are capable of representing the features of the problem without considering any specific order of the input sequence. 
In \cite{Khalil2017}, the authors propose a GNN architecture, which automatically learns policies for several graph problems such as Maximum Cut, Minimum Vertex Cover and TSP. They implement a greedy meta-algorithm design where solutions are constructed by adding graph nodes (items) sequentially to the solutions based on the graph structure to satisfy the constraints of the problem.

A widely used alternative to GNNs is the attention mechanism, the main component of the well-known Transformer architecture \cite{Vaswani2017}, which is able to selectively focus on segments of the input vector, bringing the capability of processing large sized inputs. Deudon et al. \cite{Deudon2018} and Kool et al. \cite{Kool2018} trained an architecture based on attention mechanisms improving previously reported results \cite{Vinyals2015, Bello2016, Khalil2017}.

Joshi et al. \cite{Joshi2020} analyse the generalization of these architectures to instances larger than those seen in training. The authors stated that, based on their findings, NCO methods are not robust enough to handle real world situations beyond what they see in training and do not reliably scale to practical sizes. 
As we will see in what follows, apart from the generalization issue reported by Joshi et al. \cite{Joshi2020}, the application of NCO in COPs poses a number of questions that are worth investigating. For that purpose, in this paper we present a broad analysis addressing those concerns.

\section{Critical Analysis}
\label{analysis}
As shown above, NCO models represent a breakthrough in the combinatorial optimization field, beating conventional heuristic constructive algorithms (in terms of solution quality), and exact methods (in terms of computational cost) \cite{Bello2016, Deudon2018}, by learning a policy from scratch. However, from the viewpoint of a combinatorial optimization practitioner, there are a number of questions that remain unanswered. The main difference between the application of conventional methods and DL models comes from the optimization pipeline. The general pipeline followed in a conventional optimization process starts with an (a set of) instance(s) to be solved, and a computational budget. Depending on the problem class and the budget, an algorithm (or many) is selected along with its hyper-parameters. Subsequently, the algorithm starts the optimization process and once the time expires, a result (solution to the problem instance) is provided. Conversely, DL, and consequently NCO, brings a different pipeline. After tuning the algorithm so that it meets the problem constraints, there is a training process that calibrates the parameters of a model. Once the model is trained, it can be used repeatedly to perform inference at a very low computational cost. Placing both approaches side-by-side, it is easy to see that NCO introduces an optimization pipeline that clashes with the conventional framework in a number of aspects. In what follows, we address the unanswered questions and group them in four interrelated points that need to be studied in order to fairly compare these models with conventional algorithms and contribute to the scientific progress: (1) the evaluation of the performance of the algorithm regarding the quality of the solution, (2) the training instances and the model's transferability to other instances, (3) the computational cost of the algorithm and (4) the generalization of models to larger instance sizes. 

\subsection{Performance Analysis}
When conducting an experimental comparison of an NCO method with a Conventional Optimization Algorithms (COA), the aim, generally, is to identify the best performing algorithm according to the quality of the solution. Recent works on NCO using enhanced RL methods computed for several hours were applied to the TSP, claiming to improve heuristics such as Random Insertion and Farthest Insertion\cite{Kool2018}. However, as seen in \cite{Kool2018}, there are exact methods \cite{Applegate2006} that can solve large size instances in only a few minutes. Similarly, Pan et al. \cite{Pan2021} propose an RL framework that tackles the permutation flow shop scheduling problem. The authors compare the proposal with classical heuristics and an improved versions of these. Even though these methods are useful as a baseline, they do not reflect the state-of-the-art, which relies on sophisticated hybrid metaheuristics\cite{Santucci2014}. In fact, a bibliographic comparison suggests that results obtained by NCO are less competitive than the current state-of-the-art methods, which casts doubt on the real utility of NCO.

Making a fair comparison between NCO and COA is not trivial, as the experimental setup used in both paradigms differs. Traditionally, two different stopping criteria are used for comparison: a limited computation time or a fixed number of objective-value evaluations, each of them having their supporters and detractors. In fact, most of the COAs have the ability to improve results if a larger budget is available, making it also difficult to establish a fair enough limit for all the algorithms included in a comparison. 
However, reporting the real objective values obtained by each of the proposals in the comparison, although not using the same budgets, seems to be a rigorous way to conduct the performance analysis. Moreover, comparing the NCO approach with the current state-of-the-art is a must, not with the purpose of invalidating the NCO algorithm, but to put it into perspective.

\subsection{Training Data \& Transferability}
In the optimization field, COAs are tested on different functions and/or instances of a given problem. In order to measure and compare their performance, common testbeds are required. Unfortunately, as real-world problems (instances) are difficult to obtain, most of the comparisons are made using benchmarks of artificial instances available online.



Looking at the literature, most of the works in the NCO area use randomly generated instances for training, following a common trend in combinatorial optimization. For example, in the case of TSP a grid of cities is generated by sampling uniformly at random in the unit square \cite{Bello2016,Deudon2018,Joshi2020}. Given that instance benchmarks are generally synthetic data, and are rarely examples of real problems, it seems reasonable to compare COAs and NCOs by using instance generators of the desired type, and this generator should be used in both training and validation of the algorithms.

In the following, we distinguish two strategies that are interesting when applied to train NCO models:
\begin{enumerate}
  \item \textbf{Train using instances from an instance generator}. As mentioned above, generators can create a large set of instances. These can be divided in two types, based on the instances they generate. The most common generators use uniform distributions to generate instances. Conversely, the second type of generators aim to replicate a specific type of instances, such as instances from a specific real-world problem. This could be achieved by either using a known distribution, or, although not trivial, sampling from real-world instances.
  \item \textbf{Train using a real-world benchmark}. As the training of NCO models requires large amounts of instances, this setup is limited to the case in which previously known real instances are within reach.
\end{enumerate}

Whenever random generators are used for training, it is quite common to include an intensification phase called active search\cite{Bello2016}, where the rewards obtained from the evaluation dataset are used to further tune the parameters of the pre-trained model. However, in a practical case, performing this procedure in a converged network can be time and memory intensive \cite{Hottung2021}. 

All in all, a model that learns a policy needs to be able to transfer the obtained knowledge to new instances, and in this aspect the training strategy has a big influence. The ability of random generators to draw instances from the desired target distribution is mainly what conditions the transferability of the model. Namely, uniform-distribution random generators are simple to create but they may lack transferability to real problem instances, which can not be correctly defined by simple distributions. Besides that, models trained with small sets or models with an active search learning phase are more likely to overfit and, therefore, limit their transferability. Therefore, NCO practitioners need to find a balance between diverse and large instance sets given by random generators and smaller sets that include or are closer to the real instances.




\subsection{Computational Cost \& Scalability}


NCO models need to be trained for several epochs, and once the training of the model has properly converged, they are able to infer a large number of instances in a short period of time. Conversely, COAs face each instance individually and start the optimization procedure from scratch, without any knowledge transference from instance to instance. With the increase in the number of instances to be solved, the impact of the training time in the total computation time decreases, whilst the saved time due to parallelization increases. It is not clear whether the training time needs to be considered in the computational cost comparison or not, as it would depend on how often the model has to be updated (re-trained). Even though the current literature obviates it\cite{Bello2016,Kool2018}, providing training times gives an intuition about how costly obtaining the observed performance is.

Another issue when comparing the computational efficiency of the available algorithms comes from the different programming languages used to code them and on the hardware in which they run. While COAs are generally written in C/C++ and deployed in CPUs, NCO models are mostly implemented in Python and use libraries optimized to carry out parallelized training and inference on GPUs. In order to perform a fair comparison, we should implement them in the same programming language and try to run them on similar hardware, which is neither natural nor efficient. So it seems reasonable to compare the algorithms implemented and executed in the programming languages and hardware infrastructure which the final practitioner will have easy access to, without the requirement of large overheads.

Concerning the affordability of optimization algorithms, it has been broadly reported that the use of exact methods is intractable with very large NP-hard instances, as the required computation time grows exponentially. Similarly, there is a fairly high memory/computation cost when training NCO models, which grows with both the training batch and the instance size. For this reason, an analysis on the time and memory affordability of the algorithm can give an intuition of how feasible it is to solve a certain instance.

\subsection{Generalization to different instance sizes}

The generalization capability of a model shows how the increase (or decrease) in the instance size influences the performance of models trained with smaller (or larger) instances. In general, the larger the instance size, the harder it is for a NCO model to solve it. However, most of the works in the field only reported results on instances with up to 100 nodes \cite{Bello2016, Kool2018, Deudon2018}. This may not be enough in a problem such as TSP, where the optimal solution of a problem with 100 cities can be found in 0.22 seconds using exact algorithms \cite{Bello2016, Applegate2006}. In fact, in the TSP, which has become the main playground of NCO algorithms, solving larger instances is necessary in order to see whether NCO methods are able to improve their performance with respect to exact algorithms.

Nevertheless, it is senseless to talk about generalization when working with fixed input-size models, such as sequence-to-sequence models\cite{Vinyals2015}, which can only be used for the instance size they have been trained for. These models are not very efficient, since a training period is required every time a new instance size needs to be solved. Except for the particular cases where the task consists of solving many equally sized instances, fixed input models should be replaced by other architectures or strategies that enable the computation of variable instance sizes. 

Even though GNNs solve this problem, they may lack the ability to efficiently process very large graphs. In fact, a common strategy when solving graph problems is to consider only a subset of promising nodes in the computation \cite{Manchanda2019}, for example using the k-nearest neighbor graph instead of the whole graph in each iteration. This trick is valid for the TSP since the notion of a neighborhood is direct \cite{Khalil2017, Cappart2021}. However, extending such a trick to other problems may not be that easy, requiring ad-hoc designs. 


\vspace{5mm}

As we have witnessed, each of the points mentioned above may lead to profound discussions with many inter-dependencies and implications. Along with the purpose of this paper, in the following we approach a problem that, as far as we know, has not been approached using NCO models. In fact, we will illustrate the application of NCO and compare it to COAs. To do so, we propose a series of experiments and try to give an answer to the different questions that arise in the comparison of these two paradigms.

\section{Case study: Linear Ordering Problem}
\label{problem_definition}
The Linear Ordering Problem (LOP) \cite{Ceberio2015} is a classical COP. Particularly, the LOP is a permutation problem that, in 1979, was proven to be NP-hard by Garey and Johnson. Since then, and due to its applicability in fields such as machine translation \cite{Tromble2009}, economics \cite{Leontief1986}, corruption perception \cite{Achatz2006} and rankings in sports or other tournaments \cite{Anderson2021, Cameron2021}, the LOP has gained popularity and it is easy to find a wide variety of works that have dealt with it \cite{Anderson2021_2}.

\begin{figure}
    \centering
    \includegraphics[width=0.45\textwidth]{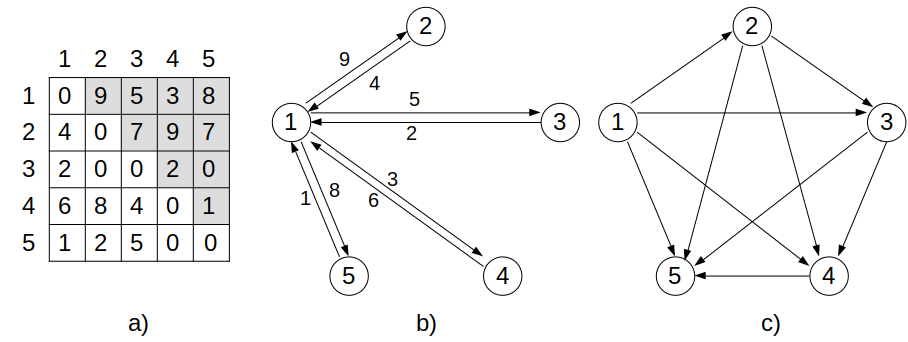}
    \caption{Example of an LOP instance of size $n=5$. \textbf{a}) The LOP instance matrix of size $n=5$ ordered as the identity permutation $\pi_e = (1 \mspace{8mu}2 \mspace{8mu}3\mspace{8mu} 4 \mspace{8mu}5)$. Entries of the matrix contributing to the objective function are highlighted in grey, the sum of the entries in the upper diagonal gives the objective value, which is 51. For this instance, the optimal solution is given by the permutation $\pi = (4 \mspace{8mu}1 \mspace{8mu}2 \mspace{8mu}5 \mspace{8mu}3)$ with a objective value of 60. \textbf{b}) Equivalent complete graph with edge weights, only the edges of the first node are shown for clarity. \textbf{c}) Identity permutation incorporated into the solution graph.}
    \label{fig:lop}
\end{figure}

Given a matrix $B = [b_{i j}]_{n \times n}$, the goal in the LOP is to find a simultaneous permutation of rows and columns such that the sum of the upper triangle entries is maximized (see Fig. \ref{fig:lop}-a). The objective function is defined formally as in Eq. \eqref{eq:lop_eqn} where $\pi$ represents the permutation that simultaneously re-orders rows and columns of the original matrix and $n$ is the problem size.
\begin{equation} 
\label{eq:lop_eqn}
  f(\pi) = \sum_{i=1}^{n-1} \sum_{j=i+1}^{n} b_{\pi_i \pi_j}
\end{equation}

An alternative formalization of the LOP is to define it as a graph problem. Let $D_n = (V_n, E_n)$ denote the complete digraph of $n$ nodes, where for every pair of nodes $i$ and $j$ there is a directed edge $(i, j)$ from $i$ to $j$ and a directed edge $(j, i)$ from $j$ to $i$, (see Fig. \ref{fig:lop}-b). 
A tournament $T$ in $E_n$ consists of a subset of edges containing for every pair of nodes $i$ and $j$ one of their directed edges. The LOP can be formulated as the problem of finding the acyclic tournament which corresponds to a linear ordering where the node ranked first is the one without incoming edges in T, the second node is the one with one incoming edge (from the node ranked first) and so on. The node ranked last is the one without outgoing edges in T, (see Fig. \ref{fig:lop}-c). The objective of the graph problem is to find an acyclic tournament that gives a ranking of the nodes that maximizes $\sum_{(i,j) \in T} c_{ij}$, where $c_{ij}$ is the weight of the directed edge $(i, j)$. 

That is,
\begin{equation}
\label{eq:linear_prog}
\begin{aligned}
\max \quad & \sum_{(i,j) \in E_n} b_{ij}x_{ij}\\
\textrm{s.t.} \quad & x_{ij} + x_{ji} = 1, \quad \textrm{for all} \; i, j \in V_n, i<j\\
  &x_{ij} + x_{jk} + x_{ki} \leq 2 \quad  i<j, i<k, j\neq k\\
  &x_{ij} \in \{0, 1\},  \quad \textrm{for all} \; i, j \in V_n  \\
\end{aligned}
\end{equation}


\section{A NCO model for the LOP}
\label{architecture}
As mentioned in the introduction, we have designed an NCO model for the LOP, so that we can run illustrative experiments which will enrich the discussion about the aspects identified in Section \ref{analysis}. Particularly, we present an autoregressive end-to-end model that returns, for each item of the LOP, the probability to be chosen (added to the solution) at each step. So, starting from an empty solution, iteratively, the model is asked for an item (the one with the highest probability) until a complete solution is obtained.

Of course, this is the most straightforward way for applying this model, but it could be used in many other ways: the model could be combined with other meta-heuristics, such as a local search, departing from the final solution provided by the model or a population-based meta-heuristic, whose initial population is obtained by sampling the output probability vector. However, the goal of this paper is to analyse the behavior, advantages and disadvantages of pure end-to-end models, and thus, the model is applied in its basic form, consequently showing a limited performance.   

\begin{figure*}
    \centering
    \includegraphics[width=0.82\textwidth]{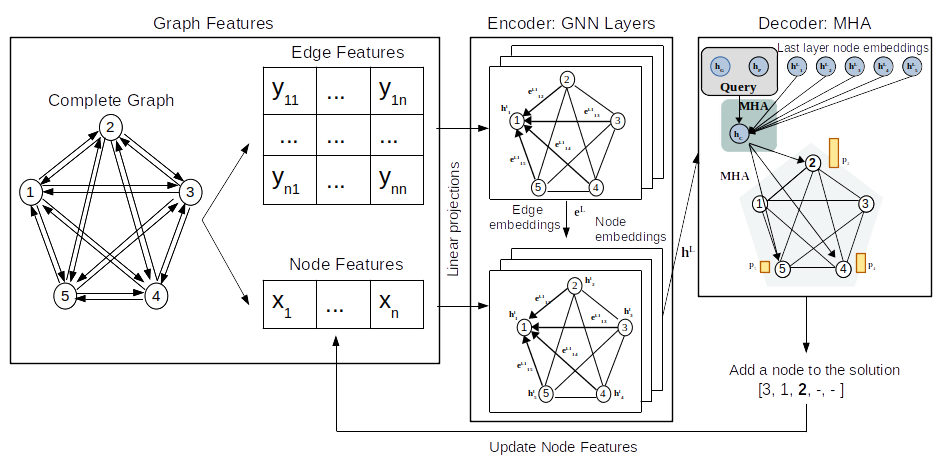}
    \caption{The NN model architecture, composed of an encoder (Message passing GNN) and a decoder (Multi-Head-Attention mechanism).}
    \label{fig:gnn}
\end{figure*}

The proposed model has two main modules: an encoder and a decoder. The problem (instance to be solved) will be represented as a graph, and the encoder will extract information from that graph, particularly node and edge features, generating high dimensional vectors called node- and edge-embeddings. Next, those embeddings are transferred into a decoder that will output the mentioned probability vector. At each step, node features will be updated according to the item(s) already placed in the previous iteration(s) until the complete permutation (solution) is obtained (see Fig. \ref{fig:gnn}).



\subsection{Graph Features}
Graph features, that is node- and edge-features are used to provide information to the model about the problem to solve, together with the actual state (placed item(s) in the solution). For example, in routing problems such as TSP, node features are used to reflect the absolute position of a city, while edge features represent the distance between two cities. Unfortunately, unlike in TSP, in the case of LOP there does not appear to be information related to a particular node, and therefore instance information will be provided only via edge-features (edge weights). Node-features will be used to codify the actual state.

Given an LOP instance matrix of size $n$, $b_{ij}$ and $b_{ji}$ represent the weights of the directed edges between the pair of nodes $i$ and $j$, i.e., their relative information. Note that $b_{ij}$ contributes to the objective value when node $i$ appears before $j$ in the permutation. On the contrary, if $j$ is placed before $i$, the term contributing to the objective or fitness value is $b_{ji}$. Therefore, we will use the difference between both values as a one-dimensional compact edge feature: $y_{ij} = - y_{ji} = b_{ij} - b_{ji}$. Edge features will be disposed as a matrix that gathers pairwise precedence information for every pair of nodes (see Fig. \ref{fig:gnn} - Graph Features).

As mentioned before, the model architecture is composed of an encoder and a decoder. A lineal projection of node and edge features forms node- and edge- embeddings, and these are fed to the encoder, which has $L$ layers. In each layer, each node gathers information from the rest of the nodes via their connection edges forming node embeddings, while edges gather information from the connected nodes, producing edge embeddings in a simultaneous way (see Fig. \ref{fig:gnn} - Encoder). Then, in a second step, embeddings are inserted into the decoder, which applies an attention mechanism (explained later) to produce the probabilities of selecting each node and appending it to the partial solution. Finally, the feature of the selected node is updated and the process repeats again (see Fig. \ref{fig:gnn} - Decoder).

\subsection{Encoder} 
Based on previous references \cite{Joshi2020}, we decided to use a message passing Graph Neural Network as an encoder. GNNs gather node ($x_i$) and edge ($y_{ij}$) features from the graph (previous step), and those feature vectors are projected linearly to produce \textit{d}-dimensional representations called embeddings. The linear projections are shown in Eq. \ref{eq:embeddings}, where $A_x$ $\in \mathbb{R}^{2 \times d}$; $A_y$, $B_x$ and $B_y$ $\in \mathbb{R}^{1 \times d}$ are learnable parameters, and $h_i^{l=1}$ and $e_{ij}^{l=1}$ denote the node and edge embeddings of the first layer ($l=1$) respectively\footnote{Note that the learnable parameters are not dependent of the instance size ($n$), instead, the learned parameters are reused $n$ times for the node projection and $n\times n$ times for the edge projection, making the model input-size invariant.}. 
\begin{equation}
\label{eq:embeddings}
    \begin{split}
        h_i^{l=1} = x_i^T * A_x  + B_x \\
        e_{ij}^{l=1} = y_{ij}^T * A_y + B_y \\
    \end{split}
\end{equation}

The encoding process consists of several message-passing neural network layers. The first layer takes node $h_i^{l=1}$ and edge $e_{ij}^{l=1}$ embeddings. In each layer, information of neighboring nodes is aggregated and, therefore, in a GNN of \textit{L} layers, the features of neighbors \textit{L} hops away are taken into account for each node.

The node $h_{i}$ and edge $e_{ij}$ embeddings at layer $l$ are defined using an $anisotropic$ message passing scheme as in \cite{Joshi2020}:
\begin{equation}
\label{eq:node_feat}
  h_{i}^{l+1} = h_{i}^{l} + ReLU\left(BN\left(W_1^l h_i^l + \sum_{j \in \mathbb{N}_i} (\sigma(e^l_{ij}) \odot W_2^l h_j^l)\right)\right)
\end{equation}
\begin{equation}
\label{eq:edge_feat}
  e_{ij}^{l+1} = e_{ij}^{l} + ReLU\left(BN\left(W_3^l e_{ij}^l + W_4^l h_{i}^l + W_5^l h_{j}^l\right)\right)
\end{equation}
where $W_1^l$, $W_2^l$, $W_3^l$, $W_4^l$ and $W_5^l$ $\in \mathbb{R}^{d \times d}$ are learnable parameters, $BN$ denotes the batch normalization layer, $\sigma$ is the sigmoid function, $\odot$ is the Hadamard product and $\mathbb{N}_i$ is the neighborhood of node $i$. In the case of a fully connected graph, as in the LOP, the neighborhood consists of every other node in the graph.

Node embeddings in the last layer $h^L$ are combined to produce the general graph representation (Eq. \eqref{eq:graph_embed}). We follow a common practice, taking the mean value over the node representations.
\begin{equation}
\label{eq:graph_embed}
 h_G = \frac{1}{n} \sum_{i=1}^n h_i^L
\end{equation}

\subsection{Decoder} 
The decoder produces the probability values that are used to take a decision about the next item to place in the partial solution. To that end, the node embeddings of the last layer and the graph representation from Eq. \eqref{eq:graph_embed} are provided to the decoder. Those node embeddings form a context vector named Query (in Fig. \ref{fig:gnn} - Decoder), that is used by an attention mechanism \cite{Kool2018} in order to obtain a probability distribution over the set of items.

The attention mechanism is a weighted message passing process where the message values acquired from the neighbors are weighted with the compatibility between the node query and the key of the neighbor. Each query vector ($Q$) is matched against a set of keys ($K$) using the dot product to measure the compatibility. In this case, the keys are the node embeddings of the last encoding layer. As noted in \cite{Vaswani2017}, having multiple attention heads ($M=8$ is suggested) allows nodes to receive different messages from different neighbors, and this strategy called Multi-Head Attention mechanism (MHA), turned out to be beneficial. 

In order to build the mentioned context vector, or query, we concatenate the graph embeddings $h_G$ from Eq. \ref{eq:graph_embed} and the embeddings of the already placed nodes. This can be seen in Eq. \ref{eq:context1}, where $[ , ]$ denotes the concatenation operation and $h_{P} = \frac{1}{n_{placed}} \sum_{i \in \pi} h_i^L$ is the aggregation of the already placed node embeddings. 
\begin{equation}
\label{eq:context1}
 \hat{h}_t^c = W_c [h_G, h_{P}]
\end{equation}

The context vector $\hat{h}_t^c$ gives additional intuition about the current state of the solution. Eq. \eqref{eq:context2} shows the query (Q), Keys (K) and Values (V) used in the MHA.
\begin{equation}
\label{eq:context2}
 h_t^c= \mathrm{MHA}(Q=\hat{h}_t^c, K =\{h_1^L, ..., h_n^L\}, V =\{h_1^L, ..., h_n^L\})
\end{equation}

Finally, a second attention mechanism, between the refined context $h_t^c$ and node embeddings $h_i^L$, produces the logits $u^c_{j}$ of the non-placed nodes:
\begin{equation}
 \label{eq:logits}
 u^c_{j} = \left\{
       \begin{array}{ll}
	 C \cdot \mathrm{tanh} \left( \frac{(W_Q h_t^c)^T \cdot (W_K h_j^L)}{\sqrt{d}} \right)      & \mathrm{if\ } j \neq \pi_{t'}\;\;  \forall t' < t \\
	 - \infty & \mathrm{otherwise} \\
       \end{array}
     \right.
\end{equation}
where the $\mathrm{tanh}$ function is used to maintain the logits within $[-C, C]$ ($C=10$). The logits at the current step $t$ are normalized using the Softmax function to produce the probabilities $p_{i}$ used to select the next item $i$ to place in the partial solution:
\begin{equation}
 \label{eq:probs}
 p_{i} = \frac{e^{u^c_{i}}}{\sum_j e^{u^c_{j}}}
\end{equation}

\subsection{Learning} 
The model is trained via the REINFORCE algorithm \cite{Williams1992}. Given an instance $s$, the output of the model with weights $\theta$, is a probability distribution $p_\theta (\pi | s)$. The training is performed minimizing the loss function
\begin{equation}
\label{eq:loss}
 \mathcal{L}(\theta | s) = \mathbb{E}_{p_\theta (\pi | s)} [-(R(\pi) - b(s)) \log p_\theta (\pi | s)]
\end{equation}
by gradient descent, where  $R(\pi) = f(\pi)$ is the reward function, which in this case is equal to the objective value of the LOP instance given a solution $\pi$, and $b(s)$ is a baseline value which is used to reduce gradient variance and increase learning speed. In order to produce the baseline, we make use of a method called self-critical sequence training (SCST) \cite{Rennie2017}. We make the model greedy, by making it take only actions with maximum probability, and then use the resulting reward as the baseline. As a result, only samples from the model that outperform the greedy action are given positive reward.

\section{Experimentation}
\label{experiments}

In the following, we will illustrate the experimental application of the end-to-end model described in the previous section, as well as consider a number of algorithms on the LOP (including the state-of-the-art) in order to compare them. As the goal is not to propose a state-of-the-art algorithm but to address the arising questions already discussed in Section \ref{analysis}, we will conduct a set of experiments to answer each one of the aspects mentioned.

\subsection{General setting}

\textbf{Instances}. As depicted in Section \ref{analysis}, we distinguish two main strategies for obtaining the instances needed to train the models: instance generators and benchmarks. In the case of the LOP, the most evident way of creating a generator is to randomly sample each element of the matrix $B$ from a uniform distribution between $(0, 1)$. Regarding benchmarks, the LOLIB \cite{Reinelt2002} is the most commonly used LOP library, which is composed of real world instances (\textit{IO}, \textit{SGB} and \textit{XLOLIB}) and randomly generated instances which try to mimic real world data (\textit{RandB},  \textit{MB}, \textit{RandA1} and \textit{RandA2}). Both strategies will be adopted for the experiments.

\textbf{Algorithms}. Among the set of conventional algorithms to solve the LOP, we will distinguish 3 groups: exact methods, constructive heuristics and metaheuristics. In each one of the groups we selected the algorithms that compose the state-of-the-art, and the stopping criteria for each algorithm will be set in a fair way so that the real performance of each algorithm can be exploited. Among the constructive heuristics, the algorithm by Becker \cite{Becker1967} is the best performing one, and as a deterministic constructive algorithm, it will be run until a solution is given. Considering exact methods, \textit{SCIP}\cite{Achterberg2009} is one of the fastest non-commercial exact solvers, it will be run until the optimal solution is found with a maximum time of 12h per instance. The implementation of the constructive heuristic and exact solver has been made following the respective handbooks. Also, we consider two of the state-of-the-art metaheuristics: A Memetic Algorithm (MA) \cite{Lugo2021}; and a Variable Neighborhood Search (CD-RVNS) \cite{Santucci2020}, which are publicly available\footnote{Codes available at \url{https://github.com/sgpceurj/Precedences_LOP} and \url{https://github.com/carlossegurag/LOP_MA-EDM}}. CD-RVNS will be stopped once $1000n^2$ objective function evaluations are computed, where $n$ is the instance size, while MA will be given the same time budget as CD-RVNS. This will be enough to let the metaheuristic converge as seen in \cite{Santucci2020, Lugo2021}. 

Finally, as for NCO methods, we will include the model described in Section \ref{architecture} (named as GNN) and an active search procedure\cite{Bello2016} (GNN-AS), which consist of performing extra training steps using the evaluation set (instances to be solved).  

\textbf{Hardware}. Models are trained in four \textit{Nvidia RTX 2070} GPUs, with a cumulative memory of 32GB. The NCO algorithm is implemented in \textit{Python 3.8}, while the conventional algorithms are written in \textit{C++}. Experiments that do not need a GPU are run on a cluster of 55 nodes, each one equipped with two \textit{Intel Xeon X5650} CPUs and 64GB of memory.

\textbf{Training}. We train four different GNN models using instances of sizes $n=20, 30, 40$ and $50$. Each model is trained for 200 epochs with 100 batches per epoch and a batch-size of 128, 128, 64 and 32 instances respectively. The batch size of larger models needs to be reduced so that the GPU memory is not overloaded. For the AS procedure, the model parameters are trained for 200 additional epochs.

\subsection{Performance Analysis}
\label{performance}

This first experiment has been designed to measure the performance of the end-to-end model compared to the algorithms described in the previous section. For this purpose, a set of 1280 random instances will be created for each $n$ size (20, 30, 40 and 50) and all the algorithms will be run to solve them. Results are depicted in Table \ref{table:performance}. For $n\leq40$, the exact algorithm provides the best solution (the optimum), but from $n=50$ onwards, it can not provide any result within the budget of 12 hours. Metaheuristics are competitive (in fact, MA provides the best results for $n=50$), and the GNN model outputs good quality solutions, with a gap between 0.3\% and 0.5\%. Moreover, when combined with active search (GNN-AS) its performance increases (0.1\% - 0.2\%), but at the cost of a notably larger computational effort. 



\begin{table}[!t]
\centering
\caption{Analysis of the performance. The gap \% to the optimal or best known value.}
\label{table:performance}
\begin{tabular}{l|r|r|r|r|} 

 \hline
 Method & n=20 & n=30 & n=40 & n=50\\ [0.5ex] 
 \hline

 Exact (ILP) & 0.00\%   & 0.00\%     & 0.00\%       & -          \\ 
 Becker      & 3.40\%     & 3.44\%     & 3.35\%      & 3.27\%   \\
 MA          & 4.9e-5\%  & 2.6e-5\%   & 1.6e-4\%   & 0.00\%   \\
 CD-RVNS     & 4.2e-4\%& 7.9e-4\%  & 2.2e-3\%  & 0.014\%   \\ 
 GNN    & 0.29\%    & 0.37\%     & 0.44\%   & 0.51\%  \\ 
GNN-AS      & 0.11\%     & 0.16\%     & 0.19\%     & 0.22\%  \\
 \hline
 
\end{tabular}
\end{table}


\subsection{Computational Cost \& Scalability}

\begin{table*}[!t]
\centering
\caption{Execution times.}
\label{table:times}
\begin{tabular}{l|c|c|c|c|c|c|c|} 
 \hline
  Method & n=20 & n=30 & n=40 & n=50 & n=100 & n=200 & n=1000\\ [0.5ex] 
 \hline

 Exact (ILP)    & 0.29s     & 25.3s      & 752s   & -       & -    & -       & -    \\ 
 Becker          & 0.03s     & 0.04s      & 0.08s   & 0.12s & 0.48s & 1.9s       & 2.2m \\
 MA            & 0.09s  & 0.22s   & 0.43s   & 0.69s  & 3.4s & 18.2s       & 2.1h \\
 CD-RVNS      & 0.09s & 0.22s   & 0.43s   & 0.69s  & 3.4s & 18.2s       & 2.1h \\ 
 GNN       & 0.07s (4h)    & 0.08s (9h)    & 0.17s (14h) & 0.19s (29h) & 0.38s & 2.1s       & 3.4m \\ 
GNN-AS         & 25m (4h)    & 1.1h (9h)    & 2.8h (14h)   & 6.6h (29h) & - & -       & - \\

 \hline
 
\end{tabular}
\end{table*}

Together with the performance, practitioners must take into account the execution time required by an algorithm to obtain the solution. In optimization contexts in which time restrictions are present, this aspect is crucial. Table \ref{table:times} shows the execution times for the different algorithms and several instance sizes. The exact algorithm quickly becomes unusable, while constructive (Becker) scales successfully. The time required by the metaheuristics grows quadratically on $n$, which can be a bottleneck for larger sizes, but for the sizes tested in this work the execution time is reasonable.

Regarding GNN, we want to first focus on the training time, the most time-demanding step, which takes several hours; up to 29 hours for $n=50$, and is unaffordable for sizes larger than 50 (at least with the available hardware). In our opinion, training time is relevant only in scenarios that require the model to be updated continuously, but could be ignored if this is not the case.
Once trained, the model shows a very fast response time, just a few minutes for the largest size tested\footnote{Since the model is size-invariant, a GNN trained with $n=50$ has been used to solve $n=100$, $200$ and $1000$ sized problems. This aspect will be explained later in section \ref{scalability}}.  Finally, the GNN-AS setup, as it includes an additional training step using the instance(s) to be solved, requires an extra computational effort (from 25 minutes to 6.6 hours), which makes this approach non-competitive and unaffordable when $n>50$.

\begin{figure}
    \centering
    \includegraphics[width=0.47\textwidth]{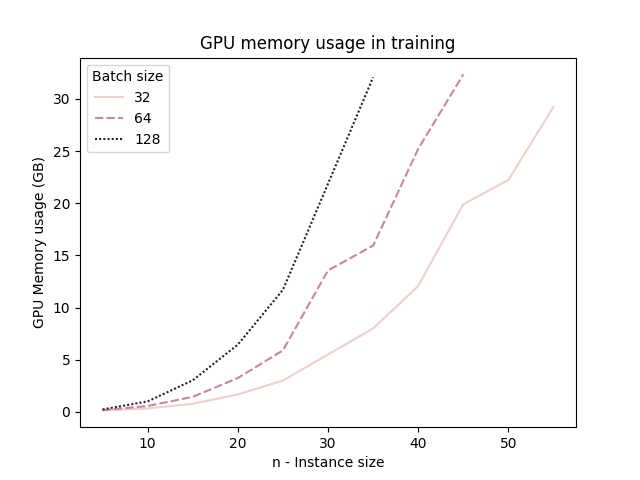}
    \caption{GPU memory use by different model sizes during training.}
    \label{fig:memory}
\end{figure}

To complete the experiment, an analysis of memory consumption has been conducted. Becker, as well as metaheuristics, follows a linear growth, which makes the memory requirements affordable. However, the memory consumption of GNNs grows polynomially with respect to the model and the batch size. Fig. \ref{fig:memory} shows the memory usage curves according to the model size, where different curves are plotted for regularly used batch sizes. It can be seen that the training phase of GNN models is really memory intensive, which quickly limits their applicability.





\subsection{Generalization to different instance sizes}
\label{scalability}

Considering the huge computational resources required by NN-based models to solve large size instances, size-invariant models are really valuable. That is, models that can be trained using different sized instances. However, even though they are usable, they may lack the ability to generalize the learnt knowledge into larger and more complex instances, guaranteeing competitive performances \cite{Joshi2020}. The definition of size-invariant models does not come from the knn strategy mentioned before for the case of the TSP, as this is an ad-hoc approach for that particular problem. In the case of the LOP, when selecting the next item, it is hard to define a subset of promising nodes, contrary to the TSP, where the proximity of the cities is tightly related to their probability to be chosen. In LOP, every item has a certain impact on the others, and therefore the whole graph needs to be considered to rank the nodes. Thus, size-invariant definition is more general, indicating that the model works node-wise (not instance-wise), and therefore it can be applied to a number of nodes (instance sizes) different from that used for training, just by iterating more times.

That being said, we conduct experiments to investigate the behaviour of the GNN model to generalize to different instance sizes. We report results provided by models trained with instances of $n=20, 30, 40$ and $50$, and evaluated in instances of $n=20, 30, 40, 50, 100, 200, 400, 700$ and $1000$. 

\begin{table*}[!t]
\centering
\caption{Performance gap to the optimal or best known value of GNN models. GNN-20 refers to the model trained with instances of size $n=20$.}
\label{table:generalization}
\begin{tabular}{l|c|c|c|c|c|c|c|c|c|} 
 \hline
 Method & n=20 & n=30 & n=40 & n=50 & n=100 & n=200 & n=400 & n=700 & n=1000\\ [0.5ex] 
 \hline
 GNN-20 & 0.29\% & 0.40\% & 0.47\% & 0.54\% & 0.78\% & 0.96\% & 0.93\% & 0.86\% & 0.79\% \\ 
 GNN-30 & 0.32\% & 0.37\% & 0.42\% & 0.48\% & 0.63\% & 0.81\% & 0.82\% & 0.77\% & 0.76\% \\ 
 GNN-40 & 0.36\% & 0.41\% & 0.44\% & 0.48\% & 0.55\% & 0.65\% & 0.71\% & 0.66\% & 0.63\% \\ 
 GNN-50 & 0.46\% & 0.46\% & 0.48\% & 0.51\% & 0.59\% & 0.64\% & 0.67\% & 0.63\% & 0.60\% \\
 \hline
\end{tabular}
\end{table*}

In the view of the results, see Table \ref{table:generalization}, the GNN model shows a good generalization capability, as the difference with the best solution worsens slightly (gaps from 0.5\% to 0.6\%). Regarding generalization, more complex models, such as GNN-40 or GNN-50 show, in general, a higher performance than the more simple ones (GNN-20 and GNN-30), which is the behaviour one would expect. In addition to the performance, it is important to remember again the quick response time of these models, an aspect that can be relevant in some scenarios.


Additionally, we compare the performance of GNNs with the rest of the proposals as a function of the instance size. Fig. 5 illustrates how the performance gap between the best performing algorithm (MA) and the GNN model increases from n = 20 to n = 200, remains constant from n = 200 to n = 400 and slightly decreases for larger sizes, which values the generalization capacity of the GNN models. Regarding the constructive algorithm (Becker), it reduces its gap with all the other algorithms consistently, even though the improvement decelerates for very large instances\footnote{Although not shown in the figure, Becker obtains a gap of 0.92\% for n = 2000}

\begin{figure}
    \centering
    \includegraphics[width=0.47\textwidth]{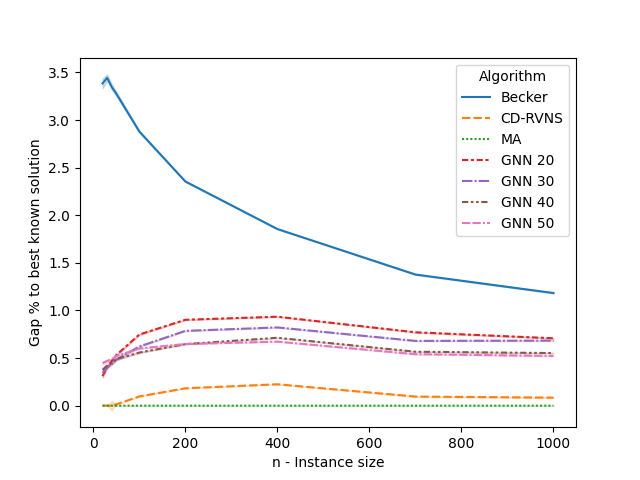}
    \caption{Gap to the best known objective value as a function of the size of instances for the analysed algorithms.}
    \label{fig:generalization}
\end{figure}

\subsection{Training Data  \& Transferability}

In order to analyse the transferability of the learnt model to other type of instances, and given that our purpose is to solve a certain target set of instances, we will consider three different training setups:

\begin{enumerate}
  \item The model is trained with random instances obtained from generators, and then target instances are solved using the model. 
  \item The model is trained in two steps, first, with random instances obtained from generators, and then an additional training (active search) is performed using the target instance or set of instances. The model will return the best solution(s) found during this active search phase. In this setup, 100 extra epochs will be dedicated to the active search.
  \item The model is trained using the target or set of target instances without any previous training.
\end{enumerate}

For the first and the second cases we will use uniformly distributed instances, and instances from the LOLIB benchmark will constitute the target set. For the third case and the active search procedure, as using directly the instances from the LOLIB is intractable due to their large size, we have created smaller instances sampling uniformly at random discrete values from the real benchmark instances in order to form a matrix (instance) of the desired size.

Table \ref{table:performance_lolib} gathers the results obtained for the different setups. Even if LOLIB instances are heterogeneous, regarding their origin or the procedure used to create them, training the model with random instances (setups 1 and 2) is notably better than training the model directly using the particular set of instances we want to solve. However, it can be observed that GNNs trained with random instance generators perform generally better in random instances (\textit{RandA1}, \textit{RandA2}, \textit{RandB} and \textit{MB}) than in real-world instances (\textit{IO}, \textit{SGB} and \textit{XLOLIB}). So, the use of generators is advisable, but these generators should be able to produce instances as close as possible to the ones we want to solve.

Finally, as seen previously, including active search is really helpful (setup 2), providing the best performing models (gaps between 0.5\% and 3.72\% with respect to the best solutions ever found).

This experiment has been also designed to give an intuition about the transferability between different LOLIB benchmarks. That is, considering the computation effort required to train a model, can this model be applied successfully to other types of instances? In this regard, interesting results have been found. For all the sets (except \textit{XLOLIB}), there are almost two models that, trained with different sets, improve the results of the model trained using the set of instances that we want to solve. Moreover, a model trained using \textit{XLOLIB} shows excellent transferability properties, obtaining the lowest gaps for all the remaining sets (except for \textit{IO}). Thus, transferability seems to be a positive characteristic of these GNN models, which should be further studied and also extended, as the whole study itself, to other problem types. 

\begin{table*}[!t]
\centering
\caption{Analysis of the performance in 7 types of instances (with different instance sizes in brackets) from the LOLIB benchmark. Gap is computed with respect to the best known value given by the Memetic Algorithm \cite{Lugo2021}. GNN-AS refers to the active search procedure applied to the model trained with instances of size 40.}
\label{table:performance_lolib}
\begin{tabular}{l |r|r|r|r|r|r|r|} 
 \hline
Method &IO (44) &  RandB (50) & SGB (75) & MB (100) & RandA1 (150) & RandA2 (200) & XLOLIB (250)\\ [0.5ex] 
 
 \hline  

 MA & 0.00\%  & 0.00\%  & 0.00\%   & 0.00\% & 0.00\%  & 0.00\%  & 0.00\% \\
 Becker & 7.14\%  & 7.49\% & 4.17\%   & 4.41\%  & 6.59\% & 1.53\%  & 7.73\% \\ 
 \hline
 (1) GNN20 & 5.21\%   & 1.27\%  & 4.80\%   & 0.54\%  & 2.47\%  & 0.93\%  & 7.78\% \\ 
 (1) GNN30 & 5.24\%   & 1.09\%  & 9.07\%   & 0.43\%  & 1.99\%  & 0.87\%  & 7.93\% \\ 
 (1) GNN40 & 5.24\%   & 1.02\%  & 6.87\%   & 0.36\%  & 1.65\% & 0.82\%  & 6.27\% \\ 
 (1) GNN50 & 5.97\%   & 1.29\%  & 10.28\%  & 0.39\%  & 1.65\%  & 1.32\% & 6.29\% \\

 \hline
 (2) GNN-AS & 0.63\%   & 0.50\%   & 2.07\%  & 0.63\%   & 1.23\%  & 0.74\%  & 3.72\% \\
 \hline
 
 (3) GNN-IO & 4.10\%   & 4.80\%  & 12.00\%  & 4.50\%  & 5.60\%  & 1.80\% & 12.00\% \\
 (3) GNN-RandB & 10.00\%   & 1.70\%  & 15.00\%  & 0.97\%  & 3.40\%  & 1.40\% & 30.00\% \\
 (3) GNN-SGB & 4.10\%   & 6.80\%  & 17.00\%  & 10.00\%  & 6.80\%  & 12.00\% & 12.00\% \\
 (3) GNN-MB & 7.50\%   & 1.50\%  & 5.60\%  & 1.50\%  & 3.10\%  & 1.50\% & 11.00\% \\
 (3) GNN-RandA1 & 5.60\%   & 2.30\%  & 13.00\%  & 1.50\%  & 3.80\%  & 1.40\% & 8.50\% \\
 (3) GNN-RandA2 & 3.40\%   & 5.00\%  & 13.00\%  & 3.60\%  & 6.20\%  & 2.30\% & 13.00\% \\
 (3) GNN-XLOLIB & 5.20\%   & 1.30\%  & 4.80\%  & 0.54\%  & 2.50\%  & 0.93\% & 7.80\% \\[1ex] 
 \hline
\end{tabular}
\end{table*}


\section{Discussion and Future Work}
\label{discussion}

Through the experimentation section, we have tested different aspects and properties of the end-to-end model, in order to analyse its behaviour and competitiveness. 

First, we observed that NCO models are not general purpose algorithms (at least not yet). Even though there are NCO models that can be suitable for a particular set of problems (e.g., GNNs for graph-based problems), they lack the capacity to adapt to all kinds of problems with the ease of metaheuristics. This can happen because we are still in the early stages of the area, and NCO development requires prior knowledge on the problem as well as advanced skills in the DL-optimization area.

We made an effort to propose a good end-to-end model for optimizing the LOP, trying to find the most competitive training strategies. Although the conducted experiments showed that the NCO model obtains a remarkable performance when compared to the constructive heuristics, it is not still able to beat state-of-the-art methods (such as MA or CD-RVNS). Not limited to that, NN-based models have a serious drawback regarding the training time and the memory requirements of larger models.

So, an obvious question arises: if metaheuristics are more competitive and easier to design, then what is the point in using NCO models? First, we are in the early stages of NN-based models for optimization, and the performance gap with respect to the state-of-the-art approaches does not seem relevant, which encourages further research on this kind of models. Secondly, looking at Table \ref{table:times}, the end-to-end model is able to provide a solution in a few minutes for the largest size ($n=1000$), while metaheurstics require a couple of hours. Thus, in environments where a fast response is required (assuming some performance loss), these models are an interesting option. For example, online decision making optimization problems (e.g., logistics). However, it must be also noted that the training time is computationally very expensive so, in order to be efficient, the model should not need to be re-trained frequently, as it would lose its competitiveness (fast response). In this regard, it is also worth noting the valuable advantage of node-wise models, such as the one designed in this work. They can be trained for $n$-sized instances and later be applied to $m >> n$-sized instances, maintaining a constant gap or even reducing it with respect to the state-of-the-art metaheuristics, which usually suffer, to a greater extent, with the increase of instance size.

Regarding the training process, there is another aspect that must be considered. Do we always need a large set of instances of the problem at hand in order to train a good-performing model? As observed in the experiments, the effectiveness of the model is greatly influenced by the quantity and the diversity of the instances used for training. While thousands of instances are required for training, existing benchmarks do not generally have enough instances. Here, random instance generators come into play. However, implementing random instance generators that produce samples with characteristics similar to the target scenarios is usually challenging. Nevertheless, we have shown that, when ad-hoc generators are not available, a successful alternative is to employ uniform random generators to train a baseline model, and if possible, incorporate active search techniques to extend the training to the particular target instance(s) to be solved. 

In summary, the experiments conducted in this work suggest that NCO models have certain capabilities that could make them a valuable choice for designing smarter metaheuristic algorithms. With this purpose in mind, recent reviews \cite{Bengio2021, Talbi2021} identify different works \cite{Hudson2021, daCosta2020, Wu2021} that employ NCO models (based on different NN variants and approaches) to choose the most adequate parameters and operators for certain metaheuristic paradigms. However, this can not be considered as a fresh new line, since other techniques, such as Bayesian Optimization, have already been used previously for designing hyper-heuristics whose parameters can be adapted to a particular problem. Analysing whether NCO provides better results compared to other techniques is still a pending matter.

In our opinion, there exists another relevant research topic which falls in line with the NCO model proposed in this article. The output of the NCO model developed for this work is a probability vector that is used to guide the construction of a solution for the LOP, i.e., deciding the item to place in the next empty position of the solution. That is, the model has been designed to make low-level decisions, and somehow it has shown quite good abilities to make the right decisions. So, including such a low-level module inside metaheuristics, although the correct answer is not always guaranteed, could be really helpful to guide them in a smarter fashion. For example, a possible way is to use NCO models to improve the search of trajectory-based metaheuristics, e.g., local search (LS) methods. These schemes usually employ quadratic size neighborhood structures that are computationally intensive to process. In such cases, given a problem instance and a solution, it would be very valuable to have a model which is able to propose the most promising neighborhood operation (or operations) to choose. Note that this is challenging since, in addition to the instance, the models must codify the solution (received as input) at which the LS algorithm is at each step. Although few, there are some works that can inspire research in this direction \cite{Chen2019}. Another research line is related to investigate the quantity and quality of the set of instances required to train the model, including their properties, characteristics, and the information and relations that the model is able to extract.

\section{Conclusion}
\label{conclusion}

In this paper we conducted a critical analysis of Neural Combinatorial Optimization algorithms and their incorporation in the conventional optimization framework. The analysis consists of 4 interrelated axes: the performance of the algorithm, the computational cost, the training instances and the generalization ability. In addition, we discuss the guidelines to facilitate the comparison of NCO approaches together with COAs in a rigorous manner. 
In order to provide some practicality to the conducted analysis, we proposed a new learning-based algorithm composed of a graph neural network and an attention mechanism. We selected the Linear Ordering Problem (LOP), and guided the reader during the process of implementation and evaluation of the proposed model. We compared the NCO method with a diverse set of algorithms, including an exact solver \cite{Achterberg2009}, a classical constructive heuristic \cite{Becker1967} and two state-of-the-art metaheuristics \cite{Lugo2021,Santucci2020}. Finally, we discussed the results, pointing out future research lines in the field of end-to-end models, which can be a promising paradigm towards the design of more efficient optimization methods.

\section*{Acknowledgments}
Andoni Irazusta Garmendia acknowledges a predoctoral grant from the Basque Government (ref. PRE\_2020\_1\_0023). This work has been partially supported by the Research Groups 2019-2021 (IT1244-19) and the Elkartek Program (KK-2020/00049, SIGZE, KK- 2021/00065) from the Basque Government, the PID2019-104933GB-10 and PID2019-106453GA-I00/AEI/10.13039/501100011033 research projects from the Spanish Ministry of Science. 

\vfill

\pagebreak

\end{document}